\documentclass[10pt,twocolumn,letterpaper]{article}

\usepackage[pagenumbers]{cvpr} 

\usepackage{graphicx}
\usepackage{amsmath}
\usepackage{amssymb}
\usepackage{booktabs}
\usepackage{overpic}
\usepackage[dvipsnames,table,xcdraw]{xcolor}

\usepackage{pifont}

\newcommand{\figref}[1]{Fig.~\ref{#1}}
\newcommand{\tabref}[1]{Tab.~\ref{#1}}
\newcommand{\eqnref}[1]{Eqn.~(\ref{#1})}
\newcommand{\secref}[1]{Sec.~\ref{#1}}
\newcommand{\myPara}[1]{\vspace{6pt}\noindent\textbf{#1}}
\newcommand{\sArt}{state-of-the-art~}

\graphicspath{{./figs/}}

\newcommand{\highlight}[1]{\textbf{\textcolor{MidnightBlue}{#1}}}

\newcommand{\addFig}[1]{}
\newcommand{\addFigs}[1]{}

\def\MyMthd{Conv2Former}

\usepackage[pagebackref,breaklinks,colorlinks,linkcolor=BrickRed,citecolor=RoyalBlue]{hyperref}

\def\cvprPaperID{757} 
\def\confName{CVPR}
\def\confYear{2023}

\begin{document}

\title{\MyMthd: A Simple Transformer-Style ConvNet for Visual Recognition}

\author{Qibin Hou$^1$ \quad Cheng-Ze Lu$^{1,2}$\thanks{Work done as an intern at ByteDance.} \quad Ming-Ming Cheng$^1$\thanks{Corresponding author.} \quad Jiashi Feng$^2$\\ 
$^1$TMCC, School of Computer Science, Nankai University\\
$^2$ByteDance, Singapore\\
{\tt\small https://github.com/HVision-NKU/Conv2Former}
}
\maketitle

\begin{abstract}
This paper does not attempt to design a \sArt method for visual recognition
but investigates a more efficient way to make use of convolutions to
encode spatial features.
By comparing the design principles of the recent 
convolutional neural networks (ConvNets) and Vision Transformers, 
we propose to simplify the self-attention by leveraging a convolutional
modulation operation.
We show that such a simple approach can better take advantage of the 
large kernels ($\geq 7\times7$) nested in convolutional layers.
We build a family of hierarchical ConvNets using the 
proposed convolutional modulation, termed \MyMthd.
Our network is simple and easy to follow.
Experiments show that our \MyMthd~outperforms existent popular ConvNets
and vision Transformers, 
like Swin Transformer and ConvNeXt in all ImageNet classification,
COCO object detection and ADE20k semantic segmentation.
%

\end{abstract}
\vspace{-10pt}

\section{Introduction}
\label{sec:intro}

The prodigious progress in visual recognition in the 2010s was mostly dedicated
to convolutional neural networks (ConvNets), 
typified by VGGNet~\cite{simonyan2014very}, 
Inception series~\cite{szegedy2015going,szegedy2016rethinking,szegedy2017inception}, 
and ResNet series~\cite{he2016deep,xie2017aggregated,pami21Res2net,zhou2020rethinking}, etc.
These recognition models mostly aggregate responses with large receptive fields 
by stacking multiple building blocks and 
adopting the pyramid network architecture but neglect 
the importance of explicitly modeling the global contextual information.
SENet series
\cite{hu2018squeeze,woo2018cbam,hou2021coordinate,bello2019attention}
break through the traditional design of CNNs and introduce 
attention-based mechanisms into CNNs to capture long-range dependencies, 
attaining surprisingly good performance.


\begin{figure}[t]
  \centering
  \includegraphics[width=0.9\linewidth]{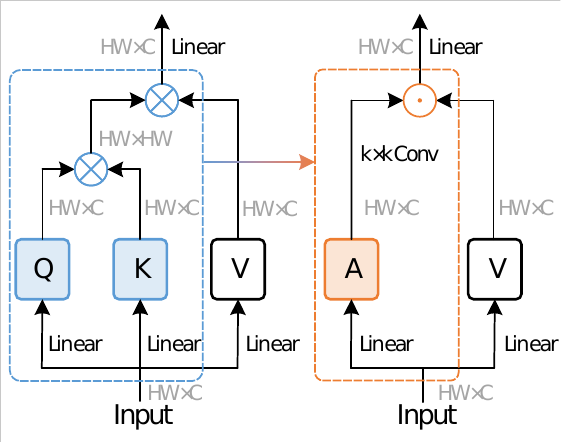}
  \caption{Comparison between the self-attention mechanism and the proposed 
    convolutional modulation operation. 
    Instead of generating attention matrices via 
    a matrix multiplication between the query and key, 
    we directly produce weights using a $k \times k$ depthwise convolution 
    to reweigh the value via the Hadamard product ($\odot$: Hadamard product; $\otimes$: matrix multiplication).
  }\label{fig:block}
\end{figure}

Since 2020, Vision Transformers (ViTs)~\cite{dosovitskiy2020image,
touvron2020training,liu2021swin,wang2021pyramid,yuan2022volo} 
further promoted the development of visual recognition models and 
show better results on the ImageNet classification and downstream tasks than
the \sArt ConvNets~\cite{tan2019efficientnet,tan2021efficientnetv2}.
This is because compared to convolutions that provide local connectivity, 
the self-attention mechanism in Transformers is able to model 
global pairwise dependencies, 
providing a more efficient way to encode spatial information 
as demonstrated in~\cite{srinivas2021bottleneck}.
Nevertheless, the computational cost caused by the self-attention 
when processing high-resolution images is considerable.

Recently, an interesting work, named ConvNeXt~\cite{liu2022convnet},
reveals that by simply modernizing the standard ResNet and 
using the similar design and training recipe as Transformers, 
ConvNets can behave even better than some
popular ViTs~\cite{liu2021swin,wang2021pyramid}.
RepLKNet~\cite{ding2022scaling} also shows the potential of leveraging
large-kernel convolutions for visual recognition.
These explorations encourage many researchers to rethink the design of ConvNets
by leveraging either 
large-kernel convolutions~\cite{guo2022visual,guo2022segnext}, 
or high-order spatial interactions \cite{rao2022hornet}, 
or sparse convolutional kernels~\cite{liu2022more}, etc.
Till now, how to more efficiently take advantage of convolutions to construct
powerful ConvNet architectures is still a hot research topic.

In this paper, we are also interested in investigating new ways to more 
efficiently make use of spatial convolutions.
Different from the ConvNeXt work~\cite{liu2022convnet} that aims to adjust the
training recipe or the position of spatial convolutions in building blocks, 
we compare the different ways ViTs and ConvNets use to 
encode spatial information.
As shown in the left part of \figref{fig:block}, 
self-attention computes the output of each pixel by a weighted summation 
of all other positions.
This process can also be mimicked by computing
the Hadamard product between the output of a large-kernel convolution and
the value representations, 
which we call convolutional modulation as depicted in the right part of 
\figref{fig:block}.
The difference is that the convolutional kernels are static while the
attention matrix generated by self-attention can adapt to the input.
Experiments show that using convolutions to generate weight matrix
yields great results as well.

Simply replacing the self-attention in ViTs with the proposed convolutional 
modulation operation yields the proposed network, termed \MyMthd.
The meaning behind it is that we aim to use convolutions to construct 
a Transformer-style ConvNet, 
in which the convolutional features are used as weights to modulate 
the value representations.
In contrast to the classic ViTs with self-attention, our method, 
like many classic ConvNets, 
is fully convolutional and hence its computations increase linearly rather 
than quadratically as in Transformers with the image resolution being higher.
This makes our method more friendly to downstream tasks, 
like object detection and high resolution semantic segmentation.
More interestingly, our \MyMthd~can benefit more from convolutions with 
larger kernels, like 11$\times$11 and 21$\times$21.
This is different from the conclusions made in previous 
ConvNets~\cite{tan2019mixconv,liu2022convnet}, 
which demonstrate using standard depthwise convolutions with kernel sizes larger than 9$\times$9 brings nearly 
no performance gain but computational burden.
We also show that our method performs better than the recent works using super large
kernel convolutions~\cite{liu2022more,ding2022scaling}.

We evaluate \MyMthd~on popular vision tasks, 
including ImageNet classification~\cite{deng2009imagenet}, 
COCO object detection/instance segmentation~\cite{lin2014microsoft}, 
and ADE20k semantic segmentation~\cite{zhou2017scene}.
To validate the capability of \MyMthd~on larger datasets, 
we also pretrain our model on the ImageNet-22k dataset and 
evaluate the performance on downstream tasks.
Experiments show that \MyMthd~performs better than popular ConvNets, 
like ConvNeXt~\cite{liu2022convnet} and 
EfficientNetV2~\cite{tan2021efficientnetv2}.
We hope our work could provide informative design choices for 
future visual recognition models.

\section{Related Work}

%
From ConvNets to the popular Vision Transformers, the architectures
have always been keeping renovating~\cite{khan2021transformers,han2022survey}.
In this section, 
we briefly describe some representative visual recognition models.

\begin{figure*}
    \centering
    \setlength{\abovecaptionskip}{2pt}
    \includegraphics[width=\linewidth]{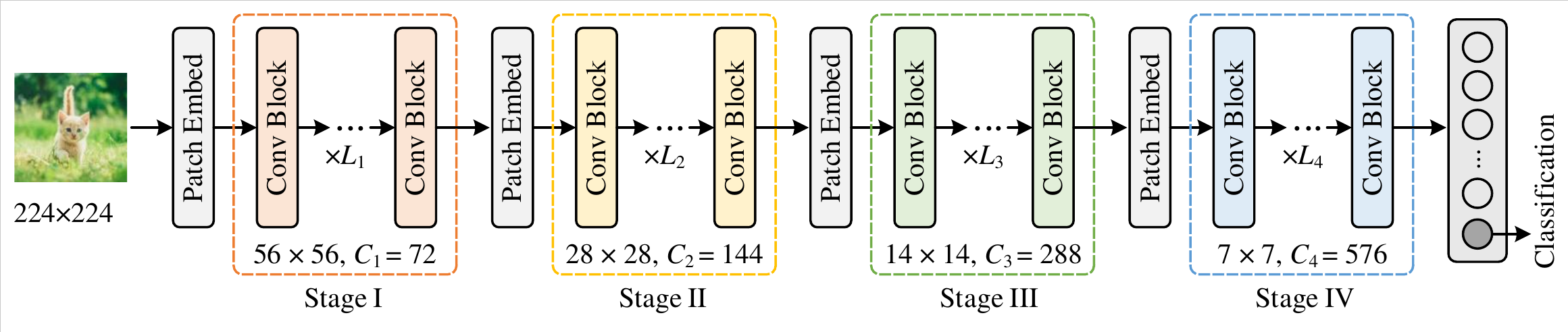}
    \caption{Overall architecture of \MyMthd. 
      Like most previous ConvNets and Swin Transformer, 
      we adopt a pyramid structure with four stages.
      In each stage, different numbers of convolutional blocks are used. 
      This figure shows the setting of the proposed \MyMthd-T, 
      where $\{L_1, L_2, L_3, L_4\} = \{3, 3, 12, 3\}$.
    }\label{fig:arch}
\end{figure*}

\subsection{Convolutional Neural Networks}

The success of early visual recognition models is mostly dedicated to the
development of ConvNets, 
typified by AlexNet~\cite{krizhevsky2012imagenet}, 
VGGNet \cite{simonyan2014very}, 
and GoogLeNet~\cite{szegedy2015going}.
These models, suffering from the gradient vanishing problem, 
mostly contain less than 20-layer convolutions.
Later, the emerging of ResNets~\cite{he2016deep} advances the conventional 
ConvNets by introducing shortcut connections, 
which make training very deep models possible.
Inceptions~\cite{szegedy2016rethinking,szegedy2017inception} and 
ResNeXt \cite{xie2017aggregated} further enrich the design principles 
of ConvNets and propose to use building blocks with multiple parallel paths of
specialized-filter convolutions.
Instead of tuning network architectures, SENet~\cite{hu2018squeeze} 
and its follow-ups~\cite{woo2018cbam,hou2021coordinate} 
aim to improve ConvNets with lightweight attention modules 
that can explicitly model the inter-dependencies among channels.
EfficientNets~\cite{tan2019efficientnet,tan2021efficientnetv2} and 
MobileNetV3~\cite{howard2019searching} take advantage of 
neural architecture search~\cite{zoph2016neural} to search for 
efficient network architectures.
RegNet~\cite{regnet} presents a new network design paradigm for ConvNets by
exploring network design spaces.

Very recently, some works aim to show the advantages of introducing 
large-kernel convolutions
\cite{ding2022scaling,rao2022hornet,guo2022visual,liu2022more,liu2022convnet}.
A typical example should be VAN~\cite{guo2022visual} that utilizes 
a depthwise convolution and a dilated one to decompose 
large-kernel convolutions.
Our \MyMthd~is different from VAN in that we do not aim to 
decompose large-kernel convolutions but show self-attention 
can be reduced to the convolutional modulation operation, which yields good recognition performance as well.
There are also some works leveraging different training 
or optimization methods or finetuning techniques
\cite{huang2019gpipe,touvron2019fixing,xie2020adversarial,brock2021high} 
to advance EfficientNet.

%

\subsection{Vision Transformers}

Transformers, originally designed for natural language processing tasks
\cite{devlin2018bert,vaswani2017attention}, 
have been widely used in visual recognition.
The most typical work should be Vision Transformer (ViT)
\cite{dosovitskiy2020image}
which shows the great potential of Transformers for processing large-scale
data in image classification.
DeiT~\cite{touvron2020training} improves the original ViT by using strong
data augmentation methods and knowledge distillation and gets rid of 
the dependence of ViTs on large-scale data.
Motivated by the success of pyramid architecture in ConvNets, 
some works~\cite{heo2021rethinking,liu2021swin,wang2021pyramid,yang2021focal,
dong2022cswin,li2022mvitv2} design pyramid structures
using Transformers to take advantage of multi-scale features.
Some works~\cite{chen2021crossvit,han2021transformer,wu2021cvt,
vaswani2021scaling,guo2022cmt,dai2021coatnet,Yuan_2021_ICCV,
han2021demystifying,hassani2022neighborhood} 
propose to introduce local dependencies into ViTs,
showing great performance in visual recognition.
Besides, there are also some works~\cite{zhou2021deepvit,zhai2021scaling,
touvron2021going,liu2022swinv2,yuan2022volo,jiang2021all} 
exploring the scaling capability of ViTs in visual recognition.
Specially, Yuan et al.~\cite{yuan2022volo} show that a two-stage ViT outperforms
the state-of-the-art CNNs on ImageNet for the first time.


\subsection{Other Models}
 
Some recent works show that mixing both Transformers and convolutions
\cite{srinivas2021bottleneck,dai2021coatnet,xu2021co}
is a promising way to develop stronger visual recognition models especially
for those aiming at efficient network design.
A typical example should be MobileViT~\cite{mehta2021mobilevit}, which
provides an efficient way to fuse both convolutions and Transformers.
EfficientViT~\cite{cai2022efficientvit}, EdgeNeXt~\cite{maaz2022edgenext},
and MobileFormer~\cite{chen2022mobile} bring back convolutions to Transformers
and show great performance in both image classification and downstream tasks.
Moreover, there are also hybrid networks that introduce different 
attention mechanisms into ConvNet for global context encoding
\cite{ramachandran2019stand,cao2020global,wang2018non,hu2018relation,hu2018gather}.
In addition, designing MLP-like architectures is also a popular research 
topic for visual recognition~\cite{tolstikhin2021mlp,touvron2021resmlp,
hou2022vision,chen2021cyclemlp,lian2021mlp,yu2022s2,rao2021global}.


\begin{table}[tp!]
  \centering
  \small
  \setlength{\tabcolsep}{1.3mm}
  \begin{tabular}{cll}
    \toprule
    Model & $\{C_1, C_2, C_3, C_4\}$ & $\{L_1, L_2, L_3, L_4\}$  \\ \midrule
    \highlight{$\bigstar$} \MyMthd-N & $\{64, 128, 256, 512\}$ & $\{2, 2, 8, 2\}$ \\ 
    \highlight{$\bigstar$} \MyMthd-T & $\{72, 144, 288, 576\}$ & $\{3, 3, 12, 3\}$ \\ 
    \highlight{$\bigstar$} \MyMthd-S & $\{72, 144, 288, 576\}$ & $\{4, 4, 32, 4\}$ \\ 
    \highlight{$\bigstar$} \MyMthd-B & $\{96, 192, 384, 768\}$ & $\{4, 4, 34, 4\}$ \\ 
    \highlight{$\bigstar$} \MyMthd-L & $\{128, 256, 512, 1024\}$ & $\{4, 4, 48, 4\}$ \\ \bottomrule
  \end{tabular}
  \vspace{-5pt}
  \caption{Brief configurations of the proposed \MyMthd. 
    We implement 5 variants with numbers of parameters 15M, 27M, 50M, 90M, 
    and 199M, respectively.
  }\label{tab:brief_conf}
\end{table}

\section{Model Design}

In this section, 
we describe the architecture of our proposed \MyMthd~and provide
some useful suggestions in model design and layer adjustment.

\subsection{Architecture}

\myPara{Overall architecture.}
The overall architecture has been shown in \figref{fig:arch}.
Similarly to the ConvNeXt~\cite{liu2022convnet} and 
Swin Transformer network~\cite{liu2021swin}, 
our \MyMthd~also adopts a pyramid architecture.
There are four stages in total, 
each of which has a different feature map resolution.
Between two consecutive stages, a patch embedding block is used to reduce the resolution, which is often a $2\times2$ convolution with stride 2.
Different stages have different numbers of convolutional blocks.
We build five \MyMthd~variants, namely \MyMthd-N, \MyMthd-T, \MyMthd-S,
\MyMthd-B, \MyMthd-L.
Details are summarized in \tabref{tab:brief_conf}.

\myPara{Stage configuration.} 
When the number of learnable parameters is fixed, 
how to arrange the width and depth of the network has an impact on 
the model performance~\cite{tan2019efficientnet,brock2021high}.
The original ResNet-50 sets the number of blocks in each stage
to $(3, 4, 6, 3)$.
ConvNeXt-T changes the block numbers to $(3, 3, 9, 3)$ following the principle
used in Swin-T and uses the stage 
compute ratio of $1:1:9:1$ for larger models.
Differently, we slightly adjust the ratios as shown in \tabref{tab:brief_conf}.
We observe that for a tiny-sized model (with less than 30M parameters) 
deeper networks perform better.
A brief comparison among four different tiny-sized models can be found
in \tabref{tab:stage_comp}.

\begin{table}[tp!]
  \centering
  \small
  \setlength{\tabcolsep}{1mm}
  \begin{tabular}{lccccc}
    \toprule
    Model & Params. & FLOPs & Stage Conf. & Top-1 Acc. \\ \midrule
    ResNet-50~\cite{he2016deep} & 26M & 4.0G & 3-4-6-3 & 78.5\% \\ 
    Swin-T~\cite{liu2021swin} & 28M & 4.5G & 2-2-6-2 & 81.5\% \\ 
    ConvNeXt-T~\cite{liu2022convnet} & 29M & 4.5G & 3-3-9-3 & 82.1\%  \\ 
    \highlight{$\bigstar$} \MyMthd-N & 15M & 2.2G & 2-2-8-2 & 81.5\% \\
    \highlight{$\bigstar$} \MyMthd-T & 27M & 4.4G & 3-3-12-3 & \highlight{83.2\%} \\ \bottomrule
  \end{tabular}
  \vspace{-5pt}
  \caption{Stage comparison with three popular models. 
    Slightly adjusting the number of convolutional blocks as shown in 
    the last row improves the performance.
  }\label{tab:stage_comp}
\end{table}

\subsection{Convolutional Modulation Block}

Our convolutional block used in each stage shares a similar structure to
Transformers, which mainly contains a self-attention layer for spatial encoding
and an FFN~\cite{xie2021segformer} for channel mixing.
Differently, we replace the self-attention layer with a simple 
convolutional modulation layer.

%

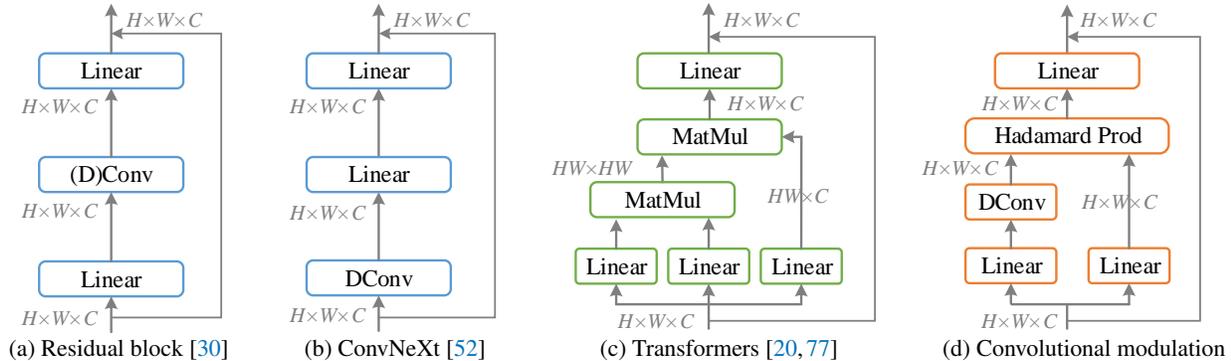
\begin{figure*}
    \centering \small
    \setlength{\abovecaptionskip}{2pt} 
    \begin{overpic}[width=0.9\linewidth]{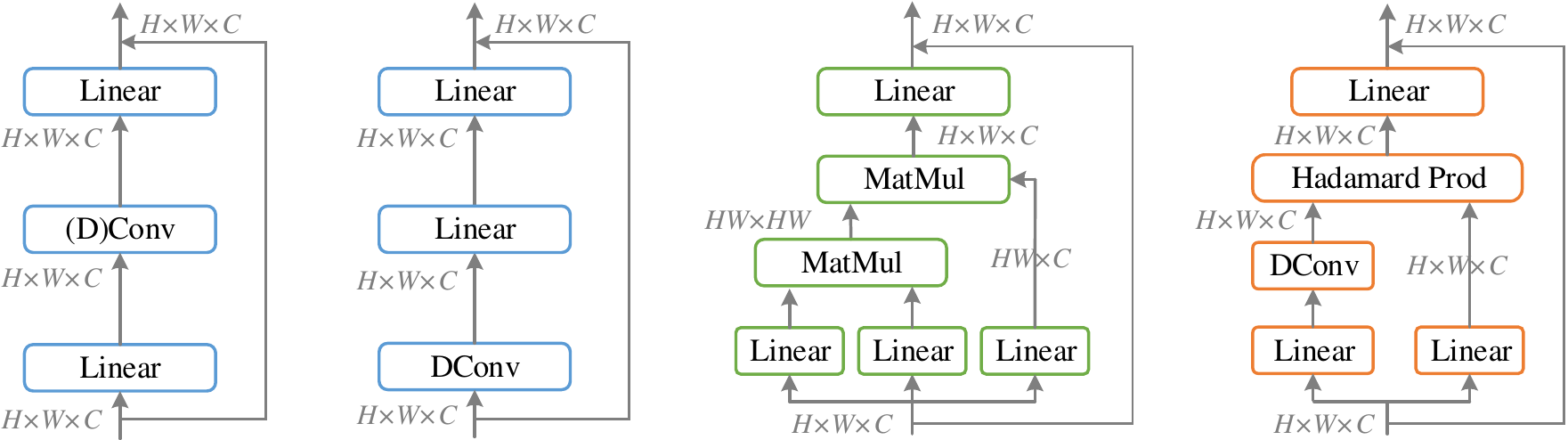}
      \put(-1,-2){(a) Residual block~\cite{he2016deep}}
      \put(24,-2){(b) ConvNeXt~\cite{liu2022convnet}}
      \put(49,-2){(c) Transformers~\cite{vaswani2017attention,dosovitskiy2020image}}
      \put(78,-2){(d) Convolutional modulation}
    \end{overpic}
    \vspace{16pt}
    \caption{Spatial encoding process comparison with self-attention 
      and typical convolutional blocks. 
      Our method uses the convolutional features from the depthwise 
      convolution as weights to modulate the value representations,
      as shown in the right linear layer in (d).
    }\label{fig:diag_comp}
\end{figure*}

\myPara{Self-attention.}
For an input token sequence $\mathbf{X}$ of length $N$,
self-attention first generates the key $\mathbf{K}$, query $\mathbf{Q}$, 
and value $\mathbf{V}$ using linear layers, where 
$\mathbf{X}, \mathbf{K}, \mathbf{Q}, \mathbf{V} \in\mathbb{R}^{N \times C}$,
$N = H \times W$, $C$ is the channel number, 
$H$ and $W$ are the spatial size of the input.
The output is the weighted average of the value
based on a similarity score $\mathbf{A}$,
\begin{equation}
    \label{eqn:sa}
    \mathrm{Attention}(\mathbf{X}) =  \mathbf{A}\mathbf{V},
\end{equation}
where $\mathbf{A}$ measures the relationships between each pair of input tokens,
which can be written as
\begin{equation}
    \label{eqn:attention}
    \mathbf{\mathbf{A}} = \mathrm{Softmax}(\mathbf{Q} \mathbf{K}^\top).
\end{equation}
Note that we omit the scaling factor for simplicity. 
In spite of the high efficiency in encoding spatial information,
the similarity score matrix $\mathbf{A}$ has a shape of $\mathbb{R}^{N \times N}$,
making the computational complexity of self-attention grows quadratically 
as the sequence length $N$ increases.

\myPara{Convolutional modulation.}
%
In our convolutional modulation layer, 
instead of calculating the similarity score matrix $\mathbf{A}$ via \eqnref{eqn:attention}, 
we simplify self-attention
by modulating the value $\mathbf{V}$ with convolutional features.
%
%
Specifically, given the input tokens $\mathbf{X} \in \mathbb{R}^{H \times W \times C}$,
we use a simple depth-wise convolution with kernel size $k \times k$ and the 
Hadamard product to calculate the output $\mathbf{Z}$ as follows:
%
\begin{align}
    \label{eqn:conv}
    \mathbf{Z} &= \mathbf{A} \odot \mathbf{V}, \\
    \mathbf{A} &= \mathrm{DConv}_{k\times k}(\mathbf{W}_1\mathbf{X}), \\
    \mathbf{V} &= \mathbf{W}_2\mathbf{X},
\end{align}
where $\odot$ is the Hadamard product, 
$\mathbf{W}_1$ and $\mathbf{W}_2$ are weight matrices of two linear layers, 
and $\mathrm{DConv}_{k\times k}$ denotes a depthwise convolution
with kernel size $k \times k$.
%
The above convolutional modulation operation enables each spatial location $(h, w)$ to be correlated with all the pixels within the $k \times k$ square region centered at $(h, w)$.
The information interaction among channels can be achieved by the linear layers.
The output for each spatial location is the weighted sum of all the
pixels within the square region.
%

\myPara{Advantages.}
A diagrammatic comparison among the residual block, self-attention, and the
proposed modulation block can be found in \figref{fig:diag_comp}.
Compared to self-attention, our method utilizes convolutions
to build relationships, which are more memory-efficient than self-attention
especially when processing high-resolution images.
Compared to the classic residual blocks~\cite{he2016deep,liu2022convnet},  our method
can also adapt to the input content due to the modulation operation.

\subsection{Micro Design}

\myPara{Larger kernel than 7$\times$7.} 
How to make use of spatial convolutions is important for ConvNet design.
%
Since VGGNet~\cite{simonyan2014very} and ResNets~\cite{he2016deep,xie2017aggregated}, $3\times3$ convolutions have been a standard choice for building ConvNets.
Later, the emerging of depthwise separable convolution~\cite{chollet2017xception}
changes this situation.
%
ConvNeXt shows that enlarging the kernel size 
of ConvNets from 3 to 7 can improve the classification performance.
However, further increasing the kernel size nearly brings no performance gain but
computational burden without re-parameterization~\cite{ding2022scaling,ding2021repvgg}.

We argue that the reason making ConvNeXt benefit little from
larger kernel sizes than $7\times7$ is the way to use spatial convolutions.
For \MyMthd, we observe a consistent performance gain as the kernel
size increases from $5\times5$ to $21\times21$.
This phenomenon not only happens for \MyMthd-T ($82.8 \rightarrow 83.4$) 
but also holds for \MyMthd-B with 80M+ parameters ($84.1 \rightarrow 84.5$).
Considering the model efficiency, we set the kernel size to $11\times11$ by default.
%

\myPara{Weighting strategy.} As shown in \figref{fig:diag_comp}(d), 
we consider the outputs of depthwise convolutions as weights to modulate the features 
after the linear projection.
It is worth noting that we use neither activation nor normalization layers 
(e.g., Sigmoid or $L_p$ normalization) before the Hadamard product.
This is an essential factor to attain good performance.
For example, adding a Sigmoid function as done in SENet~\cite{hu2018squeeze}
decreases the performance by more than 0.5\%.
%

\begin{table}[t]
  \setlength\tabcolsep{2.5pt}
  \small
  \renewcommand{\arraystretch}{1.}
  \centering
  \begin{tabular}{lccccc} \toprule
  Model & \#Params & FLOPs & Image Size & Top-1 Acc.\\ 
  \midrule
  \highlight{$\bigstar$} \MyMthd-N & 15M & 2.2G & 224$\times$224 & \highlight{81.5\%} \\
  ResNet50-d~\cite{he2016deep,he2019bag} & 26M & 4.3G & 224$\times$224 & 79.5\% \\
  SwinT-T~\cite{liu2021swin} & 28M & 4.5G & 224$\times$224 & 81.5\% \\
  ConvNeXt-T~\cite{liu2022convnet} & 29M & 4.5G & 224$\times$224 & 82.1\% \\
  \highlight{$\bigstar$} \MyMthd-T & 27M & 4.4G & 224$\times$224 & \highlight{83.2\%} \\ \midrule
  SwinT-S~\cite{liu2021swin} & 50M & 8.7G & 224$\times$224 & 83.0\% \\
  ConvNeXt-S~\cite{liu2022convnet} & 50M & 8.7G & 224$\times$224 & 83.1\% \\
  NFNet-F0~\cite{brock2021high} & 72M & 12.4G & 256$\times$256 & 83.6\% \\
  \highlight{$\bigstar$} \MyMthd-S & 50M & 8.7G & 224$\times$224 & \highlight{84.1\%} \\ \midrule
  DeiT-B~\cite{touvron2020training} & 86M & 17.5G & 224$\times$224 & 81.8\% \\
  RegNetY-16G~\cite{regnet} & 84M & 16.0G & 224$\times$224 & 82.9\% \\
  RepLKNet-31B~\cite{ding2022scaling} & 79M & 15.3G & 224$\times$224 & 83.5\% \\
  SwinT-B~\cite{liu2021swin} & 88M & 15.4G & 224$\times$224 & 83.5\% \\
  ConvNeXt-B~\cite{liu2022convnet} & 89M & 15.4G & 224$\times$224 & 83.8\% \\
  FocalNet-B~\cite{yang2022focal} & 89M & 15.4G & 224$\times$224 & 83.9\% \\
  MOAT-2~\cite{yang2022moat} & 73M & 17.2G & 224$\times$224 & 84.2\% \\
  EffNet-B7~\cite{cai2022efficientvit} & 66M & 37.0G & 600$\times$600 & 84.3\% \\
  \highlight{$\bigstar$} \MyMthd-B & 90M & 15.9G & 224$\times$224 & \highlight{84.4\%} \\
  
  \bottomrule
  \end{tabular}
  \vspace{-5pt}
  \caption{Top-1 accuracy result comparison on ImageNet~\cite{deng2009imagenet}. 
    Compared to previous popular Transformers and ConvNets, 
    our \MyMthd~achieves a surprisingly
    good results for network variants with different model sizes.
  }\label{tab:in1k}
\end{table}

We want to stress that FocalNet~\cite{yang2022focal} adopts a similar weighting strategy
as ours but its motivation is different.
FocalNet aims to extract multi-level features via $3\times3$ depthwise convolutions and
global average pooling for hierarchical context aggregation.
Differently, we attempt to simplify the self-attention operation by leveraging simple
large kernel convolutions and investigate an efficient way to  make use of
large kernel spatial convolutions for ConvNets.
Our method is much simpler than FocalNet and experiments  demonstrate
the advantages of \MyMthd~over FocalNet.

\myPara{Normalization and activations.} For normalization layers, we follow the original
ViT and ConvNeXt and adopt the Layer Normalization~\cite{ba2016layer} instead of the
widely-used batch normalization~\cite{ioffe2015batch}.
For activation layers, we use GELU~\cite{hendrycks2016gaussian}.
We found that the combination of Layer Normalization and GELU brings 0.1\%-0.2\%
performance gain. 
%

\section{Experiments}

\subsection{Experiment Setup}

\myPara{Datasets.} We evaluate the classification performance of the proposed 
\MyMthd~on the widely-used ImageNet-1k dataset~\cite{deng2009imagenet}, 
which contains around 1.2M training images and 1,000 different categories.
We report the results on the validation set that has in total 50k images.
Like some other popular models~\cite{liu2021swin,liu2022convnet}, 
we also test the scaling ability of the proposed \MyMthd~
using the large-scale ImageNet-22k dataset for pretraining, 
which has around 14M images and 21,841 classes.
After pretraining, 
we use the ImageNet-1k dataset for finetuning and report results on
the ImageNet-1k validation set as well.

\begin{table}[t]
  \setlength\tabcolsep{2.8pt}
  \small
  \renewcommand{\arraystretch}{1}
  \centering
  \begin{tabular}{lccccc} \toprule
  Model & \#Params & FLOPs & Image Size & Top-1 Acc.\\ 
  \midrule
  ConvNeXt-S~\cite{liu2022convnet} & 50M & 8.7G & 224$\times$224 & 84.6\% \\
  \highlight{$\bigstar$} \MyMthd-S & 50M & 8.7G & 224$\times$224 & \highlight{84.9\%} \\ \midrule
  SwinT-B~\cite{liu2021swin} & 88M & 15.4G & 224$\times$224 & 85.2\% \\
  ConvNeXt-B~\cite{liu2022convnet} & 89M & 15.4G & 224$\times$224 & 85.8\% \\
  MOAT-2~\cite{yang2022moat} & 73M & 17.2G & 224$\times$224 & 86.0\% \\
  \highlight{$\bigstar$} \MyMthd-B & 90M & 15.9G & 224$\times$224 & \highlight{86.2\%} \\
  SwinT-B~\cite{liu2021swin} & 88M & 47.0G & 384$\times$384 & 86.4\% \\
  ConvNeXt-B~\cite{liu2022convnet} & 89M & 45.1G & 384$\times$384 & 86.8\% \\
  \highlight{$\bigstar$} \MyMthd-B & 90M & 46.7G & 384$\times$384 & \highlight{87.0\%} \\
  \midrule
  EffNet-V2-XL~\cite{tan2021efficientnetv2} & 208M & 94.0G & 480$\times$480 & 87.3\% \\
  SwinT-L~\cite{liu2021swin} & 197M & 34.5G & 224$\times$224 & 86.3\% \\
  ConvNeXt-L~\cite{liu2022convnet} & 198M & 34.4G & 224$\times$224 & 86.6\% \\
  MOAT-3~\cite{yang2022moat} & 190M & 44.9G & 224$\times$224 & 86.8\% \\
  \highlight{$\bigstar$} \MyMthd-L & 199M & 36.0G & 224$\times$224 & \highlight{87.0\%} \\
  SwinT-L~\cite{liu2021swin} & 197M & 104G & 384$\times$384 & 87.3\% \\
  ConvNeXt-L~\cite{liu2022convnet} & 198M & 101G & 384$\times$384 & 87.5\% \\
  CoAtNet-3~\cite{dai2021coatnet} & 168M & 107G & 384$\times$384 & 87.6\% \\
  \highlight{$\bigstar$} \MyMthd-L & 199M & 105.9G & 384$\times$384 & \highlight{87.7\%} \\
  \bottomrule
  \end{tabular}
  \vspace{-5pt}
  \caption{Top-1 accuracy results on ImageNet~\cite{deng2009imagenet} 
    with pretraining on the ImageNet-22k dataset. 
    We can observe consistent improvement compared to ConvNeXt. 
    Our \MyMthd-L also performs better than EfficientNetV2-XL and CoAtNet-3.
  }\label{tab:in22k}
\end{table}

\myPara{Training settings.}
We implement our model based on \texttt{PyTorch}~\cite{paszke2019pytorch}.
%
During training, we use the AdamW optimizer~\cite{loshchilov2017decoupled} 
with a linear learning rate scaling strategy 
$lr = \text{LR}_{\text{base}} \times \text{batch}\_\text{size}/{1024}$.
The initial learning rate $\text{LR}_{\text{base}}$ is set to 0.001 and 
weight decay rate is set to  $5\times 10^{-2}$ as suggested 
in previous work~\cite{liu2022convnet}.
Throughout the experiments on ImageNet, 
we randomly crop the image size to $224\times224$
and adopt some common data augmentation methods, 
such as MixUp~\cite{zhang2017mixup} and CutMix~\cite{yun2019cutmix}.
Stochastic Depth~\cite{huang2016deep}, 
Random Erasing~\cite{zhong2020random}, 
Label Smoothing~\cite{szegedy2016rethinking}, 
RandAug~\cite{cubuk2020randaugment}, 
and Layer Scale~\cite{touvron2021going} of initial value 1e-6 are used as well.
We train all the models for 300 epochs.
For experiments on the ImageNet-22k, 
we first pretrain our model on this dataset for 90
epochs and then finetuning on the ImageNet-1k dataset for 30 epochs following ConvNeXt~\cite{liu2022convnet}.


\subsection{Comparison with Other Methods}

We compare our \MyMthd~with some popular network architectures, including
Swin Transformer~\cite{liu2021swin},
ConvNeXt~\cite{liu2022convnet}, NFNet~\cite{brock2021high}, 
DeiT~\cite{touvron2020training}, RegNet~\cite{regnet}, 
FocalNet~\cite{yang2022focal},
EfficientNets~\cite{tan2019efficientnet,tan2021efficientnetv2}, 
CoAtNet~\cite{dai2021coatnet}, RepLKNet~\cite{ding2022scaling},
and MOAT~\cite{yang2022moat}. 
Note that some of them are hybrid models of CNNs and Transformers.

\myPara{ImageNet-1k.}
We first train our \MyMthd~on the ImageNet-1k dataset and show the results 
in \tabref{tab:in1k}.
For tiny-sized models ($<30$M), our \MyMthd~has 1.1\% and 1.7\% 
performance gains compared to ConvNeXt-T and SwinT-T, respectively.
Even our \MyMthd-N with 15M parameters and 2.2G FLOPs performs the same 
as SwinT-T with 28M parameters and 4.5G FLOPs.
For the base models, the performance gain decreases but there are still
0.6\% and 0.9\% improvement over ConvNeXt-B and SwinT-B.
Compared to other popular models, 
our \MyMthd~also perform better than those with similar model sizes.
Notably, our \MyMthd-B even behaves better than EfficientNet-B7 
(84.4\% v.s. 84.3\%), 
whose computations are two times larger than ours (37G v.s. 15G).

\myPara{ImageNet-22k.} We pretrain our \MyMthd~on the large ImageNet-22k
dataset and then finetune on the ImageNet-1k dataset.
This experiment can reflect the data scaling capability of our \MyMthd.
For all experiments, we follow the settings used in~\cite{liu2022convnet} 
to train and finetune the models.
The results have been listed in \tabref{tab:in22k}.
Compared to the different variants of ConvNeXt, our \MyMthd{}s
all perform better when the model sizes are similar.
Typically, our \MyMthd-B performs better than ConvNeXt-B and the MOAT-2
network, which consumes more computations than ours.
In addition, we can see that when finetuning on a larger resolution $384\times384$
our \MyMthd-L attains better result than hybrid models, like CoAtNet
and MOAT.
Our \MyMthd-L achieves the best result 87.7\%.

\myPara{Discussion.} 
Employing large-kernel convolutions is a straightforward way
to assist CNNs in building long-range relationships.
However, directly using large-kernel  convolutions ($>7\times7$) in 
existing CNN-based architectures makes the recognition models 
difficult to optimize~\cite{tan2019mixconv,liu2022convnet}.
Recently, there are a few works aiming to develop new techniques to 
evoke the utilization of large-kernel convolutions in CNNs.
In \tabref{tab:comp_conv}, we show the results by the recent 
\sArt ConvNets with different kernel sizes. 
We can see that without any other training techniques, 
like re-parameterization or using sparse weights, 
our \MyMthd~with kernel size $7\times7$ already performs better 
than other methods under the base model setting.
Using a larger kernel size $11\times11$ yields a better performance gain.
These results reflect the advantage of our convolutional modulation block.

\begin{figure*}[t!]
  \centering
  \setlength{\abovecaptionskip}{5pt}
  \small
  \setlength{\tabcolsep}{6mm}
  \begin{tabular}{ccc}
    \includegraphics[width=0.43\textwidth]{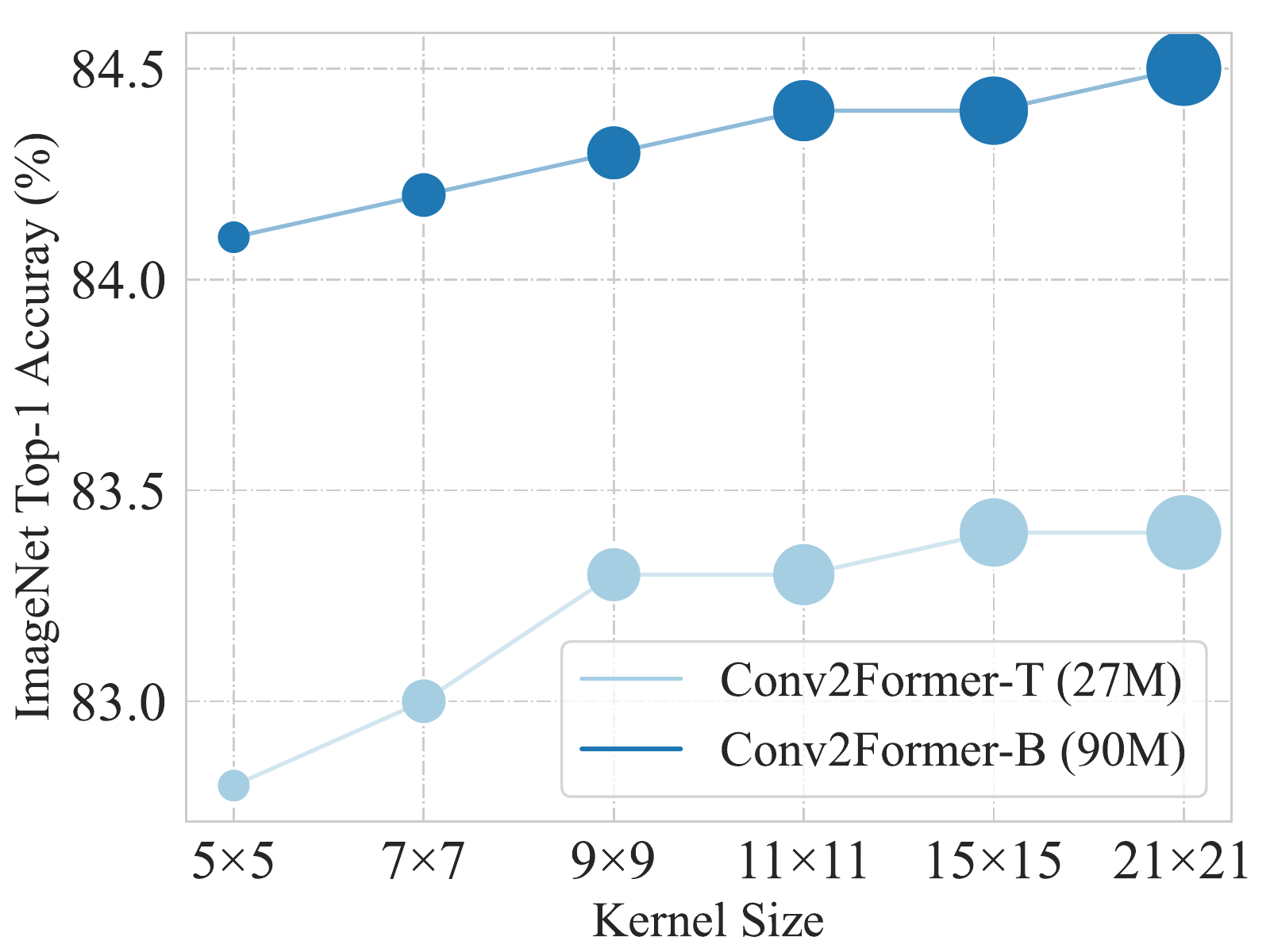} &
    \includegraphics[width=0.43\textwidth]{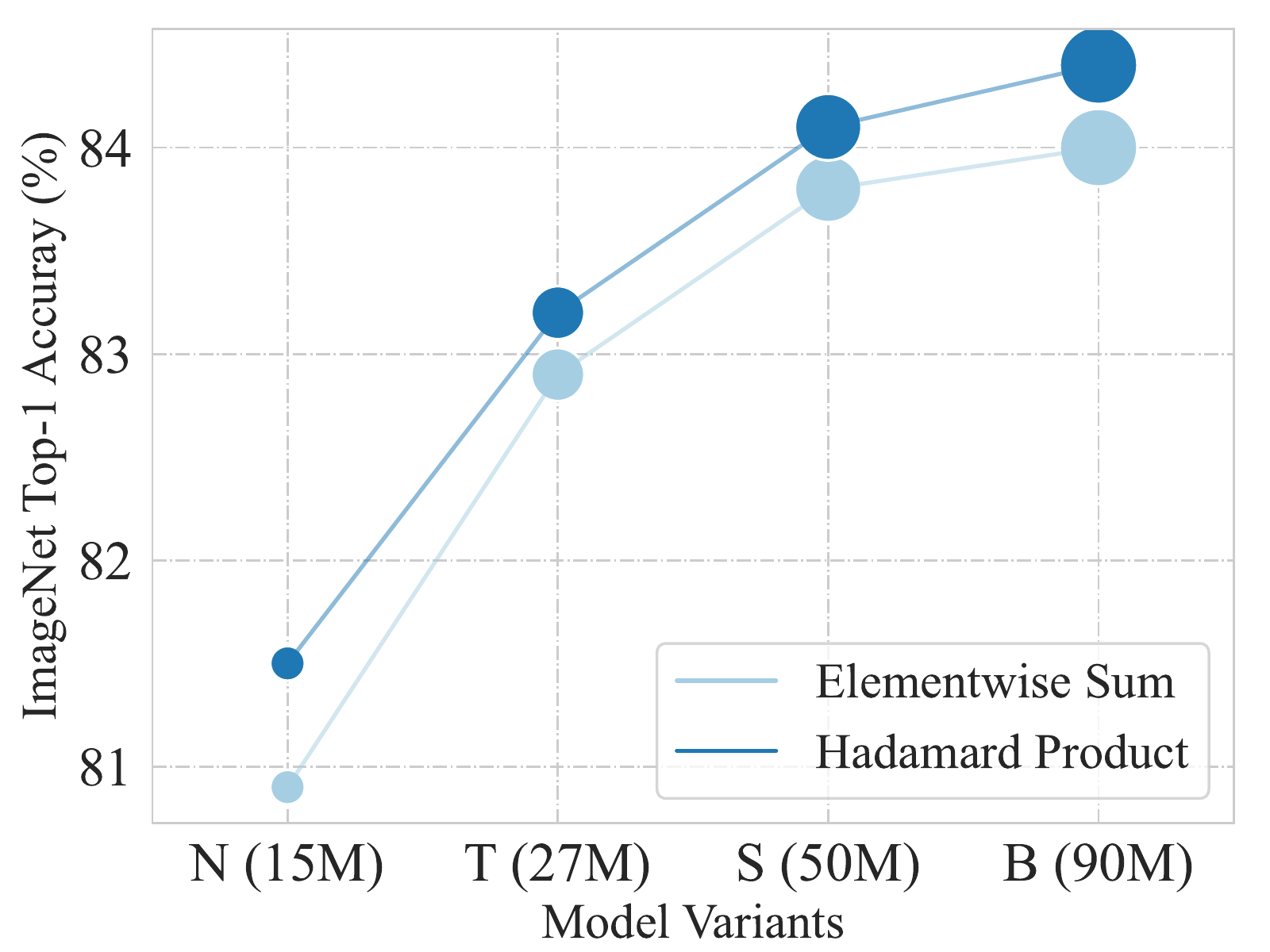} \\
    (a) Results with different kernel sizes & 
    (b) Results with different fusion strategies.\\
  \end{tabular}
  \caption{Ablative experiments. 
    For both \MyMthd-T and \MyMthd-B, 
    we can observe a consistent performance improvement when 
    increasing the kernel size from $5\times5$ to $21 \times 21$. 
    When replacing the Hadamard product with the element-wise summation operation, 
    the performance drops for all four variants of our \MyMthd.
  }\label{fig:ablation}
\end{figure*}

\begin{table}[tp]
    \setlength\tabcolsep{4pt}
    \small
    \renewcommand{\arraystretch}{1.05}
    \centering
    \begin{tabular}{lccccc} \toprule
    Model & Kernel size &\#Params & FLOPs & Acc.\\ 
    \midrule
    RepLKNet-31B~\cite{ding2022scaling} & $31\times31$ & 79M & 15.3G & 83.5\% \\
    ConvNeXt-B~\cite{liu2022convnet} & $7\times7$ & 89M & 15.4G & 83.8\% \\
    SLaK-B~\cite{liu2022more} & $51\times51$ & 95M & 17.1G & 84.0\% \\
    \highlight{$\bigstar$} \MyMthd-B & $7\times7$ & 89M & 15.6G & 84.2\% \\
    \highlight{$\bigstar$} \MyMthd-B & $11\times11$ & 90M & 15.9G & \highlight{84.4\%} \\
    \bottomrule
    \end{tabular}
    \vspace{-5pt}
    \caption{Comparison with the recent \sArt ConvNets 
      with different kernel sizes. 
      We can see that without any other training techniques, 
      like re-parameterization or using sparse weights, 
      our \MyMthd~with kernel size $11\times11$ achieves the best result. 
      These experiments indicate that our convolutional modulation operation 
      can more efficiently encode the spatial information.
      More results with different kernel sizes can be found in \secref{sec:analysis}.
    }\label{tab:comp_conv}
\end{table}

\subsection{Method Analysis} \label{sec:analysis}

In this subsection, we provide a series of method analysis on the proposed convolution modulation operation.

\myPara{Kernel size.} The ConvNeXt work~\cite{liu2022convnet} shows that
there is no performance gain when the kernel size of depthwise convolutions
is more than $7\times7$.
Here, we investigate how would the model performance change when larger
kernel sizes are used.
We select 6 different kernels for the depthwise convolutions, \emph{i.e.},
$\{5\times5, 7\times7, 9\times9, 11\times11, 15\times15, 21\times21\}$
and show the results based on two model variants, \MyMthd-T
and \MyMthd-B.
The results can be found in \figref{fig:ablation}(a).
The performance gain seems to saturates until the kernel size
is increased to $21\times21$.
This result is quite different from that made by ConvNeXt who
concludes that using larger than $7\times7$ kernels brings
no clear performance gain.
This indicates that using the convolutional features as weights
as formulated in \eqnref{eqn:conv} can more efficiently take advantage
of large kernels than traditional ways~\cite{he2016deep,liu2022convnet}.
%

\myPara{Hadamard product is better than summation.} 
As shown in \figref{fig:diag_comp}(d),
we use the convolutional features extracted by the depthwise convolutions
to modulate the weights of the right linear branch via the Hadamard product operation.
In our experiments, 
we have also attempt to leverage the element-wise summation to fuse the two branches.
\figref{fig:ablation}(b) shows the comparison results on our 
\MyMthd~at different model sizes.
The Hadamard product performs better than element-wise summation,
indicating convolutional modulation is more efficient  than 
summation in encoding spatial information.
We can also observe that small models benefit more from Hadamard product.

\myPara{Weighting strategy.} 
Other than the aforementioned two fusion strategies,
we also attempt to use other ways to fuse the feature maps, 
including adding a Sigmoid function after $\mathbf{A}$, 
applying $L_1$ normalization to $\mathbf{A}$, and linearly normalizing
the values of $\mathbf{A}$ to $(0, 1]$.
The results are summarized in \tabref{tab:opt_comp}.
We can see that the Hadamard product leads to better results 
than all other operations.
More interestingly, when adjusting the values of $\mathbf{A}$ to positive values 
using either the Sigmoid function or linear normalization to $(0, 1]$, 
the performance drops more.
This is different from the traditional attention mechanisms, 
like SE~\cite{hu2018squeeze} and CA~\cite{hou2021coordinate} 
that leverage the Sigmoid function before reweighing.
We leave this for future research.


\begin{table}[tp!]
  \centering
  \small
  \setlength{\tabcolsep}{2.5mm}
  \begin{tabular}{lccccc}
    \toprule
    Weighting Strategy & Top-1 Acc. \\ \midrule
    Element-wise sum & 82.7\% \\ 
    Adding a Sigmoid function after $\mathbf{A}$ & 82.3\%  \\ 
    Adding an $L_1$ normalization after $\mathbf{A}$  & 82.8\% \\
    Linearly normalizing the values of $\mathbf{A}$ to $(0, 1]$ & 82.2\% \\ 
    \highlight{$\bigstar$} Hadamard product &  \highlight{83.2\%} \\  \bottomrule
  \end{tabular}
  \vspace{-5pt}
  \caption{Performance comparison when different fusion strategies are used 
    in our convolutional modulation block. 
    All results are based on \MyMthd-T. 
    We can see that using the simple Hadamard product yields the best result.
  }\label{tab:opt_comp}
\end{table}

\subsection{Results on Isotropic Models to ViTs}

Different from the classic CNNs
\cite{krizhevsky2012imagenet,simonyan2014very,he2016deep} 
that adopt hierarchical architectures, 
the vanilla ViT~\cite{dosovitskiy2020image,touvron2020training} 
due to the heavy self-attention layer utilizes a plain architecture 
that contains a patch embedding layer and a stack of Transformers 
with the same sequence length.
This plain architecture has been widely used in recent works on Transformers.
Here, we follow ConvNeXt~\cite{liu2022convnet} and also attempt to 
investigate the performance of \MyMthd~
under the ViT-style architecture settings.
Similar to ConvNeXt, we set the number of blocks to 18 for both \MyMthd-IS 
and \MyMthd-IB and adjust the channel numbers to match the model size.
We use two versions of the patch embedding module: a $16\times16$ convolution
with stride 16 and three convolutions as done in~\cite{zhou2021deepvit}.

\begin{table}[tp!]
  \setlength\tabcolsep{2pt}
  \small
  \renewcommand{\arraystretch}{1}
  \centering
  \begin{tabular}{lccccc} \toprule
  Model & Patch Embed & \#Params & FLOPs & Top-1 Acc.\\ 
  \midrule
  DeiT-S & 1 Conv & 22M & 4.6G & 79.8\% \\
  ConvNeXt-IS & 1 Conv &22M & 4.3G & 79.7\% \\
  \highlight{$\bigstar$} \MyMthd-IS & 1 Conv & 23M & 4.3G & 81.2\% \\
  \highlight{$\bigstar$} \MyMthd-IS & 3 Convs & 23M & 4.5G & 82.0\% \\ \midrule
  DeiT-B & 1 Conv & 87M & 17.6G & 81.8\% \\
  ConvNeXt-IB & 1 Conv & 87M & 16.9G & 82.0\% \\
  \highlight{$\bigstar$} \MyMthd-IB & 1 Conv & 86M & 16.5G & 82.7\% \\
  \highlight{$\bigstar$} \MyMthd-IB & 3 Convs & 87M & 17.3G & 83.0\% \\
  \bottomrule
  \end{tabular}
  \vspace{-5pt}
  \caption{Comparisons among our isotropic \MyMthd, ConvNeXt, and ViT.
    `3 Convs' means that we use three convolutional layers for 
    patch embedding at the beginning of the network as done 
    in~\cite{jiang2021all,zhou2021deepvit,dai2021coatnet}. 
    For both the small-sized and base-sized models, 
    our \MyMthd~achieves better results with comparable parameters 
    and computations to other methods.
  }\label{tab:comp_iso}
\end{table}

\tabref{tab:comp_iso} shows the results.
We take the DeiT-S and DeiT-B model as baselines.
For brevity, we add a letter `I' in the model names, 
representing that the corresponding models use the isotropic architecture
as the original ViT.
We can see that for small-sized models with around 22M parameters, 
our \MyMthd-IS performs much better than DeiT-S and ConvNeXt-IS.
The performance gain is around 1.5\%.
When scaling up the model size to 80M+, 
our \MyMthd-IB achieves a top-1 accuracy score of 82.7\%, 
which is also 0.7\% better than ConvNeXt-IB and 0.9\% better than DeiT-B.
In addition, using three convolutions for patch embedding can further 
improve the result.

\subsection{Results on Downstream Tasks}
\label{sec:downstream}

In this subsection, we evaluate our method on two
downstream tasks, including object detection on COCO~\cite{lin2014microsoft}
and semantic segmentation ADE20k~\cite{zhou2017scene}.

\myPara{Results on COCO.} 
MSCOCO~\cite{lin2014microsoft} is a large dataset for object detection, 
which contains 80 categories. 
Following previous works~\cite{liu2022convnet,liu2021swin},
we conduct experiments using two popular object detectors,
Mask R-CNN~\cite{He_maskrcnn} and Cascade Mask R-CNN~\cite{cai2019cascade}
and report both the object detection and instance segmentation results.
For training, we follow the experiment settings used in 
ConvNeXt~\cite{liu2022convnet}, including multi-scale training, 
AdamW optimizer with a 3$\times$ learning schedule, 
GIoU loss~\cite{rezatofighi2019generalized}, etc.
Readers can refer to \cite{liu2022convnet,bao2021beit} for 
more detailed experimental settings.
We use the MMDetection toolbox~\cite{chen2019mmdetection} to run 
all the object detection experiments.

The results can be found in \tabref{tab:coco}.
For tiny-sized models, our \MyMthd-T achieves about 2\% AP
improvement over SwinT-T and ConvNeXt-T 
when using the Mask R-CNN framework in object detection.
For instance segmentation, the performance gain is also more than 1\%.
When using the Cascade Mask R-CNN framework, we can observe more than 1\% performance gain
than SwinT-T and ConvNeXt-T.
When scaling up the models, the improvement is also clear.


\begin{table}[tp]
    \setlength\tabcolsep{2.5pt}
    \small
    \renewcommand{\arraystretch}{1.05}
    \centering
    \begin{tabular}{lcccccc} \toprule
    Model & AP$^{\text{bb}}$ & AP$_{50}^{\text{bb}}$ & AP$_{75}^{\text{bb}}$ & AP$^{\text{mask}}$ & AP$_{50}^{\text{mask}}$ & AP$_{75}^{\text{mask}}$\\ 
    \midrule
    \multicolumn{7}{l}{\emph{Mask R-CNN~\cite{He_maskrcnn} 3$\times$ schedule}} \\ 
    SwinT-T & 46.0 & 68.1 & 50.3 & 41.6 & 65.1 & 44.9 \\
    ConvNeXt-T & 46.2 & 67.9 & 50.8 & 41.7 & 65.0 & 44.9 \\
    \highlight{$\bigstar$}  \MyMthd-T & \highlight{48.0} & 69.5 & 52.7 & \highlight{43.0} & 66.8 & 46.1 \\ \midrule
    \multicolumn{7}{l}{\emph{Cascade Mask R-CNN~\cite{cai2019cascade} 3$\times$  schedule}} \\
    SwinT-T & 50.4 & 69.2 & 54.7 & 43.7 & 66.6 & 47.3 \\
    ConvNeXt-T & 50.4 & 69.1 & 54.8 & 43.7 & 66.5 & 47.3 \\
    SLaK-T & 51.3 & 70.0 & 55.7 & 44.3 & 67.2 & 48.1 \\
    \highlight{$\bigstar$} \MyMthd-T & \highlight{51.4} & 69.8 & 55.9 & \highlight{44.5} & 67.4 & 48.3 \\
    SwinT-S & 51.9 & 70.7 & 56.3 & 45.0 & 68.2 & 48.8 \\
    ConvNeXt-S & 51.9 & 70.8 & 56.5 & 45.0 & 68.4 & 49.1 \\
    \highlight{$\bigstar$} \MyMthd-S & \highlight{52.8} & 71.4 & 57.3 & \highlight{45.7} & 69.0 & 49.8 \\
    SwinT-B & 51.9 & 70.5 & 56.4 & 45.0 & 68.1 & 48.9 \\
    ConvNeXt-B & 52.7 & 71.3 & 57.5 & \highlight{45.6} & 69.0 & 49.8 \\
    \highlight{$\bigstar$} \MyMthd-B & \highlight{52.8} & 71.1 & 57.2 & \highlight{45.6} & 68.7 & 49.3 \\ 
    \bottomrule
    \end{tabular}
    \caption{COCO~\cite{lin2014microsoft} object detection and 
      instance segmentation results using Mask R-CNN~\cite{He_maskrcnn} 
      and Cascade Mask R-CNN~\cite{cai2019cascade}.
      We report results using the ImageNet-1k dataset pretrained models.
    }\label{tab:coco}
\end{table}

\myPara{Results on ADE20k.} 
ADE20k~\cite{zhou2017scene} is a popular semantic segmentation dataset.
It contains 150 classes and a variety of scenes with 1,038 image-level labels.
Following~\cite{liu2021swin,liu2022convnet}, 
we train the models using the training set and report results 
on the validation set.
For tiny-, small-, base-sized models, 
we randomly crop the input image to $512\times512$,
and for the large-sized model, we use a crop size of $640\times640$.
We use the UperNet~\cite{xiao2018unified} as our decoder.

Results are summarized in \tabref{tab:ade20k}.
For models at different scales, our \MyMthd~can outperform
both the Swin Transformer and ConvNeXt.
Notably, there is a 1.3\% mIoU improvement compared to ConvNeXt 
at the tiny scale and the improvement is 1.1\% at the base scale.
When we further increase the model size, our \MyMthd-L with UperNet achieves
an mIoU score of 54.3\%, which is also clearly better than Swin-L and ConvNeXt-L.

\begin{table}[tp]
    \setlength\tabcolsep{2pt}
    \small
    \renewcommand{\arraystretch}{1.05}
    \centering
    \begin{tabular}{lccccc} \toprule
    Model & Pretrain & Crop Size &\#Params & mIoU (\%) \\ 
    \midrule
    SwinT-T  & ImgNet-1k & 512$^2$ & 60M & 45.8 \\
    ConvNeXt-T & ImgNet-1k & 512$^2$ & 60M & 46.7 \\
    \highlight{$\bigstar$} \MyMthd-T & ImgNet-1k & 512$^2$ & 56M  & \highlight{48.0} \\ \midrule
    SwinT-S & ImgNet-1k & 512$^2$ & 81M & 49.5 \\
    ConvNeXt-S & ImgNet-1k & 512$^2$ & 82M & 49.6 \\
    \highlight{$\bigstar$} \MyMthd-S & ImgNet-1k & 512$^2$ & 79M  & \highlight{50.3} \\ \midrule
    SwinT-B & ImgNet-1k & 512$^2$ & 121M & 49.7 \\
    ConvNeXt-B & ImgNet-1k & 512$^2$ & 122M & 49.9 \\
    \highlight{$\bigstar$} \MyMthd-B & ImgNet-1k & 512$^2$ & 120M  & \highlight{51.0} \\ \midrule
    SwinT-L & ImgNet-22k & 640$^2$ & 234M & 53.5 \\
    ConvNeXt-L & ImgNet-22k & 640$^2$ & 235M & 53.7 \\
    \highlight{$\bigstar$} \MyMthd-L & ImgNet-22k & 640$^2$ & 231M  & \highlight{54.3} \\
    \bottomrule
    \end{tabular}
    \vspace{-5pt}
    \caption{Comparisons with Swin-T and ConvNeXt on the 
      ADE20k dataset~\cite{zhou2017scene}.
      For all results, we use UperNet~\cite{xiao2018unified} as the decoder.
      At all model sizes, our \MyMthd{} achieves the best results.
    }\label{tab:ade20k}
\end{table}

\section{Conclusions and Discussions}

This paper present \MyMthd, 
a new convolutional network architecture for visual recognition.
The core of our \MyMthd~is the convolutional modulation operation that simplifies 
the self-attention mechanism by using only convolutions and Hadamard product.
We show that our convolutional modulation operation is a more efficient way 
to take advantage of large-kernel convolutions.
Our experiments in ImageNet classification, object detection, and semantic segmentation also show that our proposed \MyMthd~performs better than previous CNN-based models and most of the Transformer-based models.

\myPara{Discussion.} 
Recent \sArt visual recognition models
\cite{maaz2022edgenext,sandler2018mobilenetv2} heavily rely on convolutions 
for low-level feature encoding.
We believe there is still a large room to improve CNN-based models 
for visual recognition. 
For instance, how to more efficiently take advantage of 
large-kernel convolutions ($\ge 7\times7$), 
how to use convolutions with fixed-sized kernels to more effectively
capture large receptive fields, 
and how to more effectively introduce lightweight attention mechanisms 
to CNNs all deserve further investigation.

\myPara{Limitations.} 
This paper aims at studing how to more effectively
make use of large-kernel convolutions. 
We only pay attention to the design of CNN-based models.
%
%
How to combine the proposed convolutional modulation block 
with Transformers warrants future study.
%

%

\appendix

\section{More Experimental Settings}

\subsection{ImageNet-1k/22k}

For all experiments, we use the cosine learning rate decay schedule for training.
Most of other hyper-parameters used in our \MyMthd{} have been described in the main paper.
Compared to ConvNeXt~\cite{liu2022convnet}, we do not use layer-wise lr decay~\cite{bao2021beit} and EMA as we found they do not help in our \MyMthd{} training.
Here, we show the stochastic depth rate we use for different variants of our \MyMthd{}.
The stochastic depth rates we use for different model variants (pre)training can be found in \tabref{tab:pretrain}.

\begin{table}[htp!]
  \centering
  \small
  \setlength{\abovecaptionskip}{5pt}
  \setlength{\tabcolsep}{0.6mm}
  \begin{tabular}{cccccc} \toprule
    Model & Dataset & Stochastic Depth Rate  \\ \midrule
    \MyMthd{-N/T/S/B} & ImageNet-1k & 0.1/0.15/0.3/0.7 \\ 
    \MyMthd{-S/B/L} & ImageNet-22k & 0.1/0.2/0.1  \\
    \bottomrule
  \end{tabular}
  \caption{Stochastic depth rate for (pre)training on ImageNet-1k/22k.
  }\label{tab:pretrain}
\end{table}

Those for finetuning on ImageNet-1k can be found in \tabref{tab:finetune}.

\begin{table}[htp!]
  \centering
  \small
  \setlength{\abovecaptionskip}{5pt}
  \setlength{\tabcolsep}{2.6mm}
  \begin{tabular}{cccccc} \toprule
    Model & Resolution & Stochastic Depth Rate  \\ \midrule
    \MyMthd{-S/B/L} & 224$\times$224 & 0.2/0.2/0.3  \\ 
    \MyMthd{-B/L} & 384$\times$384 & 0.2/0.3  \\
    \bottomrule
  \end{tabular}
  \caption{Stochastic depth rate for pretraining on ImageNet22k and finetuning on ImageNet-1k.
  }\label{tab:finetune}
\end{table}

\subsection{COCO Detection}

When training on the COCO~\cite{lin2014microsoft} datasets, we following the experimental settings as in~\cite{bao2021beit,liu2022convnet}, except the stochastic depth rates which are listed in~\tabref{tab:coco_setting}.

\begin{table}[htp!]
  \centering
  \small
  \setlength{\abovecaptionskip}{5pt}
  \setlength{\tabcolsep}{1.0mm}
  \begin{tabular}{cccccc} \toprule
    Detection model  & Mask R-CNN & Cascade Mask R-CNN  \\ \midrule
    Backbone  & \MyMthd{-T} & \MyMthd{-T/S/B}  \\ 
    Stochastic Depth Rate & 0.2 & 0.2/0.6/0.9  \\ 
    \bottomrule
  \end{tabular}
  \caption{Stochastic depth rate for finetuning on COCO.
  }\label{tab:coco_setting}
\end{table}

\subsection{ADE20k Semantic Segmentation}

For semantic segmentation experiments on ADE20k~\cite{zhou2017scene}, we following the settings used in~\cite{liu2021swin} except the stochastic depth rates that are summarized in~\tabref{tab:ade20k_setting}.

\begin{table}[htp!]
  \centering
  \small
  \setlength{\abovecaptionskip}{5pt}
  \setlength{\tabcolsep}{0.6mm}
  \begin{tabular}{cccccc} \toprule
    Model & Pretraining Dataset & Stochastic Depth Rate  \\ \midrule
    \MyMthd{-T/S/B} & ImageNet-1k & 0.2/0.3/0.5  \\ 
    \MyMthd{-L} & ImageNet-22k & 0.4  \\
    \bottomrule
  \end{tabular}
  \caption{Stochastic depth rate for finetuning on Ade20k.
  }\label{tab:ade20k_setting}
\end{table}

{\small
\bibliographystyle{ieee_fullname}
\bibliography{egbib}
}

\end{document}